\documentclass[twocolumn,times,final,authoryear]{elsarticle}

\usepackage{prletters}

\usepackage{amssymb}
\usepackage{latexsym}

\setcitestyle{numbers,square}

\usepackage{kotex}
\usepackage{url}
\usepackage{xcolor}
\definecolor{newcolor}{rgb}{.8,.349,.1}

\usepackage{graphicx}
\usepackage{booktabs}

\usepackage{multirow}
\usepackage{multicol}
\usepackage{bbm}
\usepackage{color, colortbl}
\definecolor{Gray}{gray}{0.9}

\usepackage{subcaption}
\usepackage{graphicx}

\usepackage{tabulary}
\usepackage{tabularx}
\usepackage{bm}
\usepackage{makecell}
\usepackage{diagbox}

\newcommand{\etal}{\textit{et al.}}

\usepackage{url}
\usepackage{xcolor}
\definecolor{newcolor}{rgb}{.8,.349,.1}

\journal{Pattern Recognition Letters}

\begin{document}

\setcounter{page}{1}

\begin{frontmatter}

\title{SAMIRO: Spatial Attention Mutual Information Regularization with a Pre-trained Model as Oracle for Lane Detection}

\author[1]{Hyunjong \surname{Lee}}
\author[2]{Jangho \surname{Lee}\corref{cor1}} 
\author[1]{Jaekoo \surname{Lee}\corref{cor1}} 

\cortext[cor1]{ Principal corresponding author: Jaekoo Lee (jaekoo@kookmin.ac.kr) }


\affiliation[1]{organization={Kookmin University},
                addressline={77 Jeongneung-ro Seongbuk-gu}, 
                city={Seoul},
                postcode={02707}, 
                country={Republic of Korea}}
                
\affiliation[2]{organization={Incheon National University},
                addressline={119, Academy-ro, Yeonsu-gu}, 
                city={Incheon}, 
                postcode={22012}, 
                country={Republic of Korea}}

\received{1 May 2013}
\finalform{10 May 2013}
\accepted{13 May 2013}
\availableonline{15 May 2013}
\communicated{S. Sarkar}



\begin{abstract}
Lane detection is an important topic in the future mobility solutions.
%
Real-world environmental challenges such as background clutter, varying illumination, and occlusions pose significant obstacles to effective lane detection, particularly when relying on data-driven approaches that require substantial effort and cost for data collection and annotation.
%
%
%
To address these issues, lane detection methods must leverage contextual and global information from surrounding lanes and objects.
In this paper, we propose a \textit{Spatial Attention Mutual Information Regularization with a pre-trained model as an Oracle}, called \textit{SAMIRO}.
%
%
SAMIRO enhances lane detection performance by transferring knowledge from a pre-trained model while preserving domain-agnostic spatial information.
Leveraging SAMIRO's plug-and-play characteristic, we integrate it into various state-of-the-art lane detection approaches and conduct extensive experiments on major benchmarks such as CULane, Tusimple, and LLAMAS. 
%
%
%
The results demonstrate that SAMIRO consistently improves performance across different models and datasets. 
The code will be made available upon publication.
\end{abstract}

\begin{keyword}
Action recognition, artificial intelligence, deep learning, real-time system, video understanding
\end{keyword}

\end{frontmatter}


\section{Introduction}
%
Lane detection is a fundamental component that enables vehicles to perceive road conditions and detect lane markings accurately, making it an essential task for autonomous driving~\cite{c1}.
%
However, achieving accurate lane detection remains a significant challenge due to factors such as background clutter, varying illumination, and occlusions~\cite{lane_review_for_vehicle, lane_robust}.
Since inaccurate lane detection can lead to accidents, it is essential to have a reliable and robust lane perception system that operates well under various road conditions.
%

Existing state-of-the-art (SOTA) lane detection methods can be broadly categorized into two primary approaches.
%
%
The first approach focuses on designing additional modules to enhance the detection of irregular lane characteristic~\cite{scnn,resa,laneformer}.
For example, RESA~\cite{resa} introduces a propagation mechanism that passes the spatial representations across both the row and column directions of feature maps, enhancing lane feature representations. While effective, this approach requires additional modules, leading to a trade-off between bottleneck and high accuracy.

The second approach aims to improve lane detection by enhancing feature representations through improved training schemes~\cite{clld,sad}.
For instance, CLLD~\cite{clld} employs self-supervised learning (SSL) to understand various patch images, thereby enriching the learned feature representations.
This approach does not require additional modules.
However, unlike other computer vision tasks, SSL has not been sufficiently investigated for its impact on lane detection performance~\cite{clld, ssl-survey}.
Despite the recent advancements, lane detection still faces challenges, particularly in generalizing to diverse road conditions with background clutter, illumination changes, and occlusions.

\begin{figure*}[!t]
\begin{center}
\includegraphics[width=0.8\textwidth]{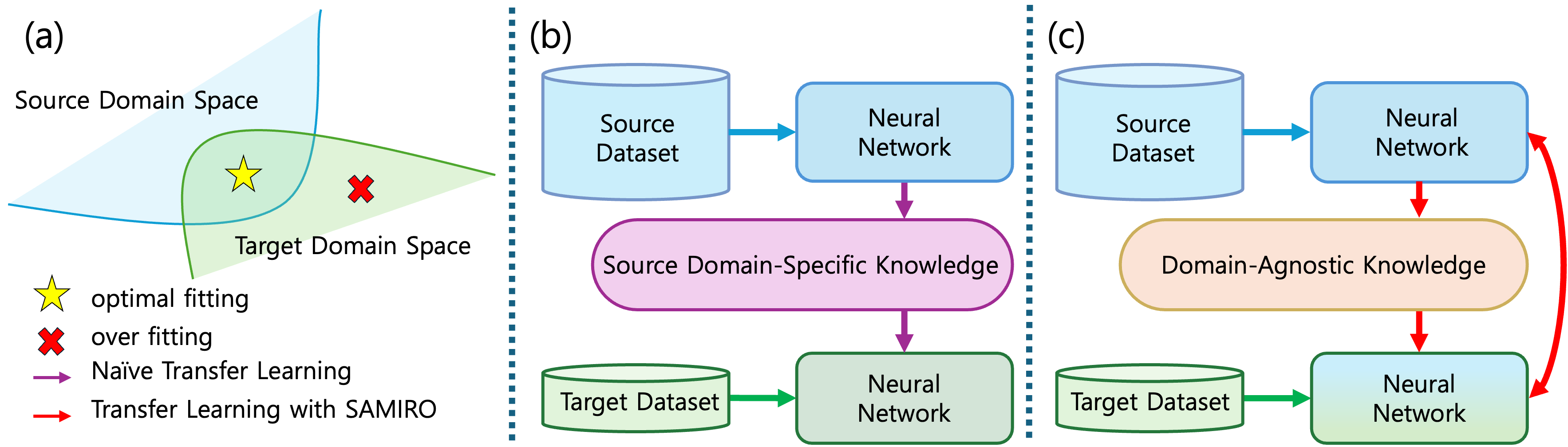}
\end{center}
   \caption{\hspace{0.01 cm} Illustration of transfer learning. (a) Describes a scenario in transfer learning. (b) Provides an overview of Naïve transfer learning, which fails to address the domain gap between source and target data. (c) Represents a scenario where continuous interaction between source and target data through SAMIRO leads to learning domain-agnostic knowledge.}
   \vspace{-1em}
\label{fig:teaser}
\end{figure*}


To address these limitations and enhance generalization in lane detection, we propose a \textit{Spatial Attention Mutual Information Regularization with a pre-trained model as an Oracle}, called \textit{SAMIRO}.
%
%
%
%
Specifically, our method is derived from a regularization method for domain generalization introduced MIRO~\cite{miro}.
SAMIRO effectively transfers knowledge from a pre-trained model, trained on abundant general image datasets (source domain), to a model trained on limited lane detection datasets (target domain).
%

As illustrated in Figure~\ref{fig:teaser}, typical transfer learning conveys domain-specific knowledge, which may not generalize well if the source and target domains have different properties and distributions.
%
%
This discrepancy can lead to overfitting and degraded performance in the target domain.
%
To address this issue in transfer learning, we propose SAMIRO, a novel framework that continuously transfers domain-agnostic knowledge from a pre-trained model to the lane detection model, thereby enhancing performance.

SAMIRO comprises two key components: (i) spatial attention, (ii) mutual information regularization with a pre-trained model as an oracle. By incorporating spatial attention, the lane detection model focus on spatially relevant features crucial for lane detection. The mutual information regularization ensures that the lane detection model learns representations similar to the domain-agnostic knowledge of the pre-trained oracle model. 
%
Through extensive experiments on benchmark datasets~\cite{scnn,tusimple,llamas}, we demonstrate that SAMIRO significantly improves the lane detection model's performance.

Our main contributions are summarized as follows: 
(i) MI‑based, domain‑agnostic spatial knowledge transfer improves robustness to illumination/occlusion/clutter.
(ii) plug‑and‑play design enables broad, low‑friction adoption.
(iii) consistent gains across CULane, TuSimple, and LLAMAS demonstrate practical value.

\vspace{-0.5em}


\section{Related Works}
\textbf{Auxiliary Learning Modules}
%
In recent lane detection studies, most approaches have focused on auxiliary learning modules to enhance lane detection by addressing lane disappearance caused by background clutter, occlusions, and perspective distortion~\cite{scnn, resa, robust_lane}.
%
Pan \etal~\cite{scnn} introduced spatial CNN (SCNN), which captures spatial relationships by extending traditional convolutions from layer-by-layer to slice-by-slice within feature maps. Similarly, Zheng \etal~\cite{resa} proposed the recurrent feature-shift aggregator (RESA) module, utilizing prior knowledge about lane contours by shifting feature maps vertically and horizontally to enrich spatial information.
%
%
While these methods effectively enhance spatial features, they often come at the cost of increased computational complexity due to pixel-wise predictions. Laneformer~\cite{laneformer} effectively integrates relationships between neighboring objects and lanes by incorporating an auxiliary detection module.

\textbf{Enhancing Feature Representation}
%
Alternatively, some approaches aim to directly enhance feature representations for lane detection without introducing additional learning modules.
Inspired by the dispersion of lane features, SAD~\cite{sad} reinforces the lane features between layers through self-attention distillation.
Peng~\etal~\cite{active} redesigned the conventional active learning by employing efficient feature selection based on knowledge distillation.
However, these methods often suffer from significant performance degradation when encountering perspective variation or distortion not seen during training.
%

\begin{figure*}[!t]
    \begin{center}
    \includegraphics[width=0.7\textwidth]{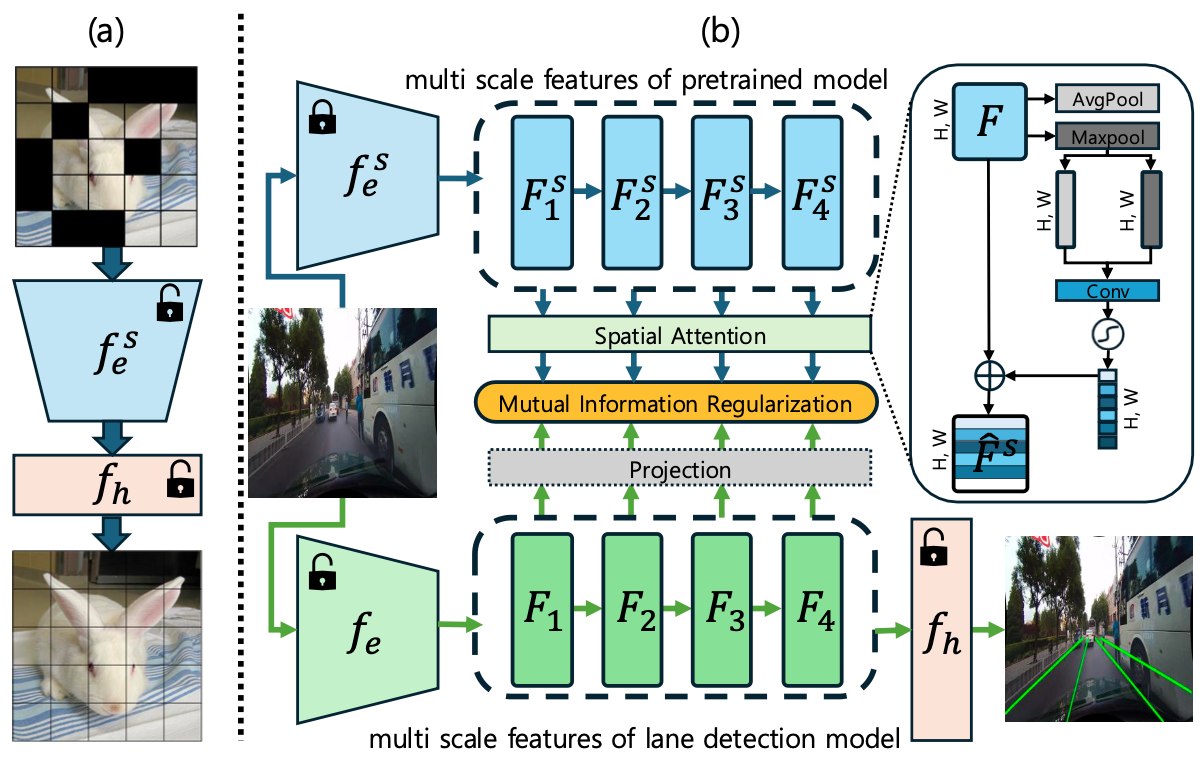}
    \end{center}
    \caption{\hspace{0.01 cm} Overall framework. (a) pre-training with masked image modeling, (b) transfer learning with SAMIRO.}
   \vspace{-2em}
    \label{fig:overview}
\end{figure*}

\textbf{Self-supervised Learning (SSL)}
%
%
SSL aims to learn intrinsic representations or relations within data without relying on external labels provided by humans~\cite{ssl_survey}.
It helps enhance generalization performance by learning the inherent properties from the data~\cite{JiGen,selfreg,self_domain1,miro}.
Kim~\etal~\cite{selfreg} successfully learned a domain-invariant representation using only positive data pairs without using the negative data pairs.
The method of masked image modeling (MIM)~\cite{mim_survey}, as proposed by Kong and Zhang~\cite{mim_robust}, has demonstrated its capability to acquire robust features against occlusion.
CLLD~\cite{clld} applies transfer learning to the lane detection dataset following MIM training on the ImageNet~\cite{imagenet} dataset.
%
However, SSL approaches generally require substantial computational resources and large-scale training data.

\textbf{Regularization}
%
%
%
The goal of regularization is to reduce the overfitting and improve the generalization performance.
Well-known methods like L1/L2 regularization~\cite{l1l2} and dropout~\cite{dropout} impose constraints on network weights.
Recently, feature-level regularization methods have gained attention~\cite{regularizing,miro}.
CS-KD~\cite{regularizing} distills knowledge from the logit distributions obtained from samples of the same class.
%
MIRO~\cite{miro} effectively enhanced domain generalization performance in the target domain through feature-level regularization using pre-trained models.
%
However, these regularization strategies are designed for limited domains and architectures. Especially, they have been explored only to a limited extent in lane detection tasks. 
\vspace{-0.5em}

\section{Method}


\subsection{Overall Framework}
As shown in Figure~\ref{fig:overview}, our proposed framework consists of two main training schemes: (a) masked image modeling (MIM) based self-supervised learning (SSL) for pre-training and (b) transfer learning with SAMIRO for fine-tuning.
First, pre-training in our proposed framework is based on MIM~\cite{spark}, a well-known SSL, which involves masking a portion of the input images and learning to predict the masked portions.
%
The MIM-based pre-trained is conducted using masked ImageNet data, which are fed into the encoder $f_{e}^{s}$ and the lane detection head $f_{h}^{s}$ to efficiently learn latent representations, as described in~\cite{spark}.
Then, the knowledge from the pre-trained encoder $f_{e}^{s}$ is transferred to the encoder for lane detection $f_{e}^{t}$ using the SAMIRO.
SAMIRO improves lane detection performance by efficiently incorporating  domain-agnostic knowledge from $f_e^s$ into $f_e^t$.
%
In Figure~\ref{fig:overview} (b), the feature maps of $f_e^s$ and the corresponding feature maps of $f_e^{t}$ are defined as $F_l^s\in \mathbb{R}^{C\times H\times W}$ and $F_l^{t}\in \mathbb{R}^{C\times H\times W}$, respectively, where $C$ is the number of channels, $H$ is the height, $W$ is the width, and $l$ represents the layer index $[1, L]$.
%
%
Consequently, SAMIRO extracts domain-agnostic feature representations in a data-driven way and minimizes the distance between the features of $F^s$ and $F^t$.
Then, precise lane detection is performed by passing the fine-tuned $F^t$ to the lane detection head $f_{h}^{t}$.
\subsection{Spatial Attention Mutual Information Regularization with a pre-trained model as an Oracle (SAMIRO)}
SAMIRO is built with the following two components:
(i) \textit{spatial attention} extracts hierarchical spatial features from the pre-trained model to aid in lane detection.
%
(ii) \textit{mutual information regularization with a pre-trained model as an oracle} conveys useful knowledge in a data-driven manner.

\textbf{Spatial attention}
%
The key component of the lane detection task is extracting features robust to background clutter, illumination, and occlusion under diverse road conditions. 
Since the pre-training model is learned to predict the masked region, it can yield rich hidden representations~\cite{mim_robust}.

Effective use of these representations requires leveraging spatial information in lane detection~\cite{salane, scnn}. 
Accordingly, we propose integrating spatial attention into the SAMIRO framework. 
The spatial attention technique analyzes the spatial characteristics of the input image, guiding the model to focus on the most informative regions.
The spatial attention is derived from CBAM~\cite{cbam} and processes multi-stage feature maps to obtain two types of spatial maps, each containing compressed information across all channels through max pooling and average pooling. 
After concatenating the two spatial maps, a convolution operation and a sigmoid activation are applied to compute spatial attention weights $W_{SA}$ that emphasize important information at the spatial level. 
The $W_{SA}$ are then element-wise multiplied with the original feature maps to produce the final spatial attention values. 
The output $\hat{F_l^s}$ of the spatial attention can be expressed as follows: $W_{SA} = \sigma(p([AvgPool(F_l^s); MaxPool(F_l^s)]))$ and $\hat{F_l^s} = W_{SA} \otimes F_l^s$
%
, where $\sigma$ is the sigmoid operator, $p$ is the convolutional layer and $\otimes$ denotes the element-wise product.

\textbf{Mutual information regularization with a pre-trained model as an oracle}
We regularize the lane detection model to learn feature representations similar to the oracle knowledge from a large pre-trained model, deriving a tractable approximation of lane detection as the target objective. 
%
%
The following loss function provides an efficient approximation of the loss function in MIRO~\cite{miro}:
%
$L_{MIRO} = \frac{1}{N}\sum\limits_{i=1}^N (\log {\vert \mathbf{w_c} \vert} + \Vert F_{i, l}^{s} - F_{i, l}^{t} \Vert_{\mathbf{w_c}^{-1}}^{2})$
, where $\mathbf{w_c} \in \mathbb{R}^{C\times1\times1}$ is a learnable channel-wise scaling parameter, and $N$ is defined as $C \cdot H \cdot W$.
$L_{MIRO}$ improves computational efficiency by replacing the mutual information loss with mean squared error for each channel and learning the inter-channel balance from the data. 
%
However, $L_{MIRO}$ often suffers from training instability issues.
To address this issue, we introduce the $\mathrm{ReLU}$ function to the $\mathrm{log}$ regularization term.
This modification helps ensure the stability of the loss term and contributes to reducing instability during the model training process.
In addition, we utilize an MIM-based pre-trained model as an oracle.
However, due to discrepancies in architecture between the pre-trained model and the target model, the scale of the feature maps may vary, potentially leading to performance degradation during the knowledge transfer process~\cite{feature}.

To address this issue, we propose a loss function for minimizing the scale discrepancy between the two models to transfer the normalized features.
The loss function can be expressed mathematically as follows: $L_{SAMIRO} = \frac{1}{N} \sum\limits_{i=1}^N (\mathrm{ReLU} (\log {\vert \mathbf{w_c}\vert}) + \Vert \frac{\hat{F_{i, l}^{s}}}{ \Vert \hat{F_{l}^{s}} \Vert} - \frac{g(F_{i, l}^{t})}{ \Vert F_{l}^{t} \Vert}\Vert_{\mathbf{w_c}^{-1}}^{2})$.
%
%
, where $\hat{\Vert F_{l}^{s}\Vert}$ and $ \Vert F_{l}^{t} \Vert$ are the normalized feature maps following the channel axis. and $g$ is a projection composed of a convolutional layer (channel‑wise linear/1×1 projection), which serves to align the intermediate features $\hat{F^s_{i,l}}$ and $F^t_{i,l}$

\textbf{Total loss function and overall training scheme}
The total loss $L$ is defined as the combination of lane detection loss $L_{LD}$~\cite{resa, ufsa, clrnet, clrernet, bezier, laneatt} and our regularization loss $L_{SAMIRO}$: $L=L_{LD} + \lambda L_{SAMIRO}$.

\begin{table*}[!t]
    \scriptsize
    \caption{ Experimental results on CULane dataset. Bold text represents the best results.\label{tab:table1}}
    \centering
    \begin{tabular}{c|c|ccccccccccc|}
    
        \hline
        
        Method & Backbone & F1 ($\uparrow$) & Normal & Crowded & Dazzle & Shadow & No line & Arrow & Curve & Cross & Night\\
        
        \hline
        
        RESA~\cite{resa} & ResNet34~\cite{resnet} & 74.50 & 91.90 & 72.40 & 66.50 & 72.00 & 46.30 & 88.10 & 68.60 & 1896 & 69.80 \\
        RESA\textbf{+Ours}~\cite{resa} & ResNet34~\cite{resnet} & 75.19 & 92.12 & 72.64 & 65.36 & 72.66 & 48.78 & 88.16 & 68.98 & 1304 & 69.51 \\
        RESA~\cite{resa} & ResNet50~\cite{resnet} & 75.30 & 92.10 & 73.10 & 69.20 & 72.80 & 47.70 & 88.30 & 70.30 & 1503 & 69.90 \\
        RESA\textbf{+Ours}~\cite{resa} & ResNet50~\cite{resnet} & 75.84 & 92.70 & 74.03 & 68.12 & 75.31 & 48.19 & 88.65 & 70.97 & 1583 & 70.36 \\
        
        \hline \hline
        
        UFLD~\cite{ufsa} & ResNet34~\cite{resnet} & 70.51 & 89.44 & 68.06 & 58.49 & 65.34 & 41.37 & 85.11 & 59.29 & 1947 & 65.54 \\
        UFLD\textbf{+Ours}~\cite{ufsa} & ResNet34~\cite{resnet} & 71.25 & 90.01 & 69.60 & 58.65 & 62.48 & 42.41 & 85.55 & 58.91 & 1811 & 65.39 \\
        UFLD~\cite{ufsa} & ResNet50~\cite{resnet} & 70.88 & 89.57 & 69.04 & 60.25 & 68.71 & 41.44 & 83.73 & 58.89 & 1732 & 64.90 \\
        UFLD\textbf{+Ours}~\cite{ufsa} & ResNet50~\cite{resnet} & 71.34 & 89.93 & 69.25 & 61.41 & 66.82 & 42.20 & 84.78 & 61.45 &1580 & 65.81 \\
        
        \hline \hline
        
        CLRNet~\cite{clrnet} & ResNet34~\cite{resnet} & 79.73 & 93.49 & 78.06 & 74.57 & 79.92 & 54.01 & 90.59 & 72.77 & 1216 & 75.02 \\
        CLRNet\textbf{+Ours}~\cite{clrnet} & ResNet34~\cite{resnet} & 79.83 & 93.47 & 78.79 & 75.53 & 80.71 & 52.91 & 90.03 & 71.24 & 1175 & 75.26 \\
        CLRNet ~\cite{clrnet} & ResNet50~\cite{resnet} & 79.75 & 93.58 & 78.07 & 71.70 & 81.74 & 53.58 & 89.79 & 72.29 & 1135 & 74.91 \\
        CLRNet\textbf{+Ours}~\cite{clrnet} & ResNet50~\cite{resnet} & 80.02 & 93.75 & 78.77 & 72.84 & 80.75 & 53.88 & 90.07 & 70.74 & 1507 & 75.02 \\
        
        \hline \hline
        
        CLRerNet~\cite{clrernet} & ResNet34~\cite{resnet} & 80.76 & 93.93 & 79.51 & 73.88 & 83.16 & 55.55 & 90.87 & 74.45 & \textbf{1088} & 76.02 \\
        CLRerNet\textbf{+Ours}~\cite{clrernet} & ResNet34~\cite{resnet} & 81.07 & 94.21 & 79.91 & 75.17 & 83.53 & 56.55 & \textbf{91.37} & 79.18 & 1476 & 76.59 \\
        CLRerNet ~\cite{clrernet} & ResNet50~\cite{resnet} & 81.30 & 94.19 & 80.13 & 73.84 & 83.67 & 56.30 & 90.81 & 78.90 & 1221 & 76.96 \\
        CLRerNet\textbf{+Ours}~\cite{clrernet} & ResNet50~\cite{resnet} & 81.45 & 94.28 & 80.79 & 74.37 & 84.39 & 57.03 & 90.39 & 79.12 & 1357 & 76.61 \\
        CLRerNet~\cite{clrernet} & DLA34~\cite{dla34} & 81.43 & 94.36 & 80.62 & 75.23 & 84.35 & 57.31 & 91.17 & 79.11 & 1540 & 76.92 \\
        CLRerNet\textbf{+Ours}~\cite{clrernet} & DLA34~\cite{dla34} & \textbf{81.64} & \textbf{94.42} & \textbf{80.81} & \textbf{75.83} & \textbf{84.48} & \textbf{57.55} & 90.57 & \textbf{80.25} & 1434 & \textbf{77.07} \\
        \hline
    \end{tabular}
\end{table*}
\vspace{-0.5em}

\section{Experiments}

\subsection{Datasets}
%
%
In this paper, the effectiveness of SAMIRO was demonstrated using lane recognition benchmark datasets CULane~\cite{scnn}, Tusimple~\cite{tusimple}, and LLAMAS~\cite{llamas}.

\textbf{CULane}~\cite{scnn} consists of images captured from cameras mounted on vehicles driving in urban areas, comprising $88,800$ training images, $9,675$ validation images, and $34,680$ testing images.
The testing data includes various environments across nine different scenarios, such as Normal, Crowded, Night, No line, Shadow, Arrow, Dazzle light, Curve, and Crossroad.
\textbf{Tusimple}~\cite{tusimple} was collected from cameras mounted on highway vehicles.
It contains $3,626$ training images and $2,782$ testing images.
Contrary to the CULane dataset, it was primarily collected on highways containing multiple lanes under clear visibility.
%
%
\textbf{LLAMAS}~\cite{llamas} is a large-scale lane detection dataset consisting of $100,000$ images.
Since the labels for the test set are not publicly available, this paper reports the detection results on the validation set.


\subsection{Evaluation and Implementation Details}
%
%
To evaluate lane detection performance on CULane and LLAMAS, the F1 score was used as the evaluation metric.
Following the previous studies~\cite{resa,ufld2,clrnet}, we model a predicted lane line with a width of $30$ pixels as a correctly predicted lane line.
We then compute the intersection-over-union (IoU) between these predictions and the corresponding ground-truth pixels.
Here, the lane with derived IoU that exceeds a certain value is regarded as the true positive (TP).
In this study, we set a threshold of $0.5$.
Based on the predictions and the ground-truth pixels, the following two indicators are defined as follows: $\mathrm{Precision} = \frac{TP}{TP+FP}$, $\mathrm{Recall} = \frac{TP}{TP+FN}$
%
%
with two evaluation measures, the F1 is defined as follows: $\mathrm{F1} = \frac{2 \times \mathrm{Precision} \times \mathrm{Recall}}{\mathrm{Precision} \times \mathrm{Recall}}$.

%
%

In the Tusimple dataset, we adopt the accuracy as our evaluation criterion, with $C_{clip}$ representing the correctly predicted lane points and $S_{clip}$ denoting the total number of ground-truth points for each clip.
$C_{clip}$ is counted when the distance between predicted lane points and actual points is located within a certain range.
The accuracy with $C_{clip}$ and $S_{clip}$ is defined as follows: $\mathrm{Accuracy} = \frac{ \sum_{clip} C_{clip} }{ \sum_{clip} S_{clip} }$.
%
%
During the training and evaluation process on the CULane, Tusimple and LLAMAS datasets, we adjusted the image resolution following the previous studies~\cite{resa,ufsa,clrnet,clrernet,laneatt,bezier}.
Based on the plug-and-play nature of the proposed objective function, we only incorporated the SAMIRO loss in the training procedure.
All experiments were conducted using NVIDIA A6000 GPUs.
%




\begin{figure*}[!t]
    \centering
    \includegraphics[width=\textwidth]{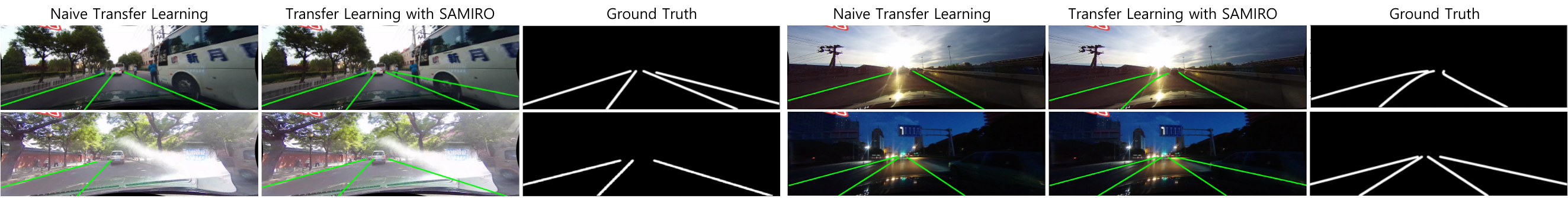}
    \caption{Visualized comparison of the CULane dataset. The base represents the results of CLRerNet. The results show challenging situations in lane recognition, such as viewpoint variation, background clutter, illumination changes, and occlusions by vehicles, reflections, sunlight, and nighttime.}
    \label{result_vis}
    \vspace{-1.5em}
\end{figure*}
\vspace{-1em}

\subsection{Results and Discussion}
%
Table~\ref{tab:table1} presents the experimental results on the CULane dataset.
Since the proposed method operates in a plug-and-play manner, it can be applied to four strong competitors: message passing-based RESA~\cite{resa}, row anchor-based UFLD~\cite{ufsa}, lane anchor-based CLRNet~\cite{clrnet}, and the state-of-the-art method CLRerNet~\cite{clrernet}.
In addition, to facilitate the application of the proposed method, we utilize the ResNet34/50~\cite{resnet} pretrained model for ResNet~\cite{resnet} backbone experiments, and the ConvNeXt-S~\cite{convnext} pre-trained model for DLA34~\cite{dla34} backbone experiments.

The experimental results demonstrate performance improvements with our method across all lane detection models.
%
%
When our method was applied to RESA~\cite{resa} with the message passing-based ResNet34~\cite{resnet} backbone, it achieved a performance gain of $0.69\%$.
In addition, applying SAMIRO to CLRerNet~\cite{clrernet} with DLA34~\cite{dla34} backbone yields a performance improvement of $0.21\%$, attaining $81.64\%$ and setting a new state-of-the-art lane detection F1 score.
These results demonstrate the effectiveness of the proposed method in robust lane detection.
As shown in Table~\ref{tab:table1}, the F1 score improves by 0.21--0.7\%, a gain that plays a crucial role in ensuring the reliability and safety of autonomous driving systems under real-world conditions.

\begin{table}[!t]
    \scriptsize
    \caption{ Experimental results on Tusimple dataset.\label{tab:table2} }
    \centering
    \begin{tabular}{c|c|ccc}
        \hline
        Method &
        Backbone & 
        ACC. ($\uparrow$) & 
        FP ($\downarrow$) & 
        FN ($\downarrow$) \\
        \hline
        RESA~\cite{resa}& ResNet34~\cite{resnet} & 96.82 & 3.63 & 2.48 \\
        RESA\textbf{+Ours}~\cite{resa} & ResNet34~\cite{resnet} & 96.82 & 2.67 & 2.57 \\
        RESA~\cite{resa} & ResNet50~\cite{resnet} & 96.83 & 3.52 & 2.64 \\
        RESA\textbf{+Ours}~\cite{resa} & ResNet50~\cite{resnet} & \textbf{96.89} & 4.31 & 2.11 \\
        \hline \hline
        UFLD~\cite{ufsa} & ResNet34~\cite{resnet} & 95.81 & 19.05 & 3.92 \\
        UFLD\textbf{+Ours}~\cite{ufsa} & ResNet34~\cite{resnet} & 95.90 & 18.81 & 3.65 \\
        UFLD~\cite{ufsa} & ResNet50~\cite{resnet} & 95.81 & 19.06 & 3.89 \\
        UFLD\textbf{+Ours}~\cite{ufsa} & ResNet50~\cite{resnet} & 95.86 & 18.93 & 3.77 \\
        \hline \hline
        CLRNet~\cite{clrnet} & ResNet34~\cite{resnet} & 96.83 & 2.22 & 2.13 \\
        CLRNet\textbf{+Ours}~\cite{clrnet} & ResNet34~\cite{resnet} & 96.86 & \textbf{2.15} & 2.19 \\
        CLRNet~\cite{clrnet} & ResNet50~\cite{resnet} & 96.63 & 2.46 & 2.63 \\
        CLRNet\textbf{+Ours}~\cite{clrnet} & ResNet50~\cite{resnet} & \textbf{96.89} & 2.84 & \textbf{1.57} \\
        \hline
    \end{tabular}
\end{table}

Table~\ref{tab:table2} summarizes the experimental results on the Tusimple dataset, where three methods including RESA~\cite{resa}, UFLD~\cite{ufld2}, CLRNet~\cite{clrnet}.
The details of the other methods are described in the previous paragraph.
In the experimental results, our method contributes to improving the lane detection performance when integrated into RESA~\cite{resa} with ResNet50~\cite{resnet}, UFLD~\cite{ufsa} with ResNet34/50 backbone~\cite{resnet}, and CLRNet~\cite{clrnet} with ResNet34/50 backbone~\cite{resnet} in terms of accuracy.
Especially, SAMIRO is plugged with CLRNet~\cite{clrnet}, we achieved an accuracy of $96.89\%$.
These quantitative results imply that the proposed method is effective in highway driving conditions and proves its competitiveness compared to other models.

Table~\ref{tab:llamas} represents the experimental results on the LLAMAS dataset.
The proposed method improved overall performance when applied to $B \acute{e}zier$, LaneATT~\cite{laneatt} and CLRNet~\cite{clrnet}.
Specifically, it achieved a $0.58\%$ performance improvement on LaneATT with the ResNet34~\cite{resnet} backbone, reaching $94.90\%$.
In Figure~\ref{result_vis}, we compared transfer learning with SAMIRO with the Naïve transfer learning method.
Each image is sampled from the test data of CULane, including challenging scenarios where lane detection is hindered by factors such as occlusion due to vehicles, lighting conditions, and nighttime.
The visualization results demonstrate qualitatively that our method exhibits the closest resemblance to the ground truth compared to the Naïve transfer learning method.
The proposed SAMIRO is especially robust to occlusion caused by vehicles, reflections, sunlight, and varying road conditions.

We have incorporated subset-specific metrics from Table~\ref{tab:table1} to provide a more detailed analysis, as shown in Figure~\ref{result_vis}. Our evaluation reveals that the proposed method consistently improves performance on challenging subsets. For instance, CLRNet-R34~\cite{clrnet} F1 score on the Night subset improved from 75.02 to 75.26, and CLRerNet-DLA34~\cite{clrernet} score increased from 76.92 to 77.07. Similarly, on the Shadow subset, RESA~\cite{resa} saw a significant gain, moving from 72.80 to 75.31, while CLRNet-R34~\cite{clrnet} score rose from 79.92 to 80.71. Although a minor fluctuation was observed for CLRNet-R50~\cite{clrnet} on the Shadow subset (from 81.74 to 80.75), the overall net F1 gains and consistent improvements across multiple challenging subsets substantiate the robustness of our approach.


\begin{table}[!t]
    \scriptsize
    \caption{Comparison on LLAMAS dataset.\label{tab:llamas}}
    \centering\resizebox{\columnwidth}{!}{
    \begin{tabular}{c|c|ccc}
        \hline
        Method & Backbone & F1 ($\uparrow$)& Precision ($\uparrow$)& Recall ($\uparrow$) \\
        \hline
        $B\acute{e}zier$~\cite{bezier} & ResNet34~\cite{resnet} & 95.83 & 96.14 & 95.52 \\
        $B\acute{e}zier$\textbf{+Ours}~\cite{bezier} & ResNet34~\cite{resnet} & 95.94 & 96.16 & 95.72 \\
        $B\acute{e}zier$~\cite{bezier} & ResNet50~\cite{resnet} & 94.99 & 95.26 & 94.72 \\
        $B\acute{e}zier$\textbf{+Ours}~\cite{bezier} & ResNet50~\cite{resnet} & 95.19 & 95.68 & 94.71 \\
        \hline \hline
        LaneATT~\cite{laneatt} & ResNet34~\cite{resnet} & 94.32 & 97.21 & 91.59 \\
        LaneATT\textbf{+Ours}~\cite{laneatt} & ResNet34~\cite{resnet} & 94.90 & 97.48 & 92.45 \\
        LaneATT~\cite{laneatt} & ResNet50~\cite{resnet} & 94.74 & \textbf{97.67} & 91.98 \\
        LaneATT\textbf{+Ours}~\cite{laneatt} & ResNet50~\cite{resnet} & 94.76 & 97.46 & 92.21 \\
        \hline
        \hline
        CLRNet~\cite{clrnet} & ResNet34~\cite{resnet} & 96.71 & 97.11 & 96.32 \\
        CLRNet\textbf{+Ours}~\cite{clrnet} & ResNet34~\cite{resnet} & 96.87 & 97.24 & 96.51 \\
        CLRNet~\cite{clrnet} & ResNet50~\cite{resnet} & 96.89 & 97.40 & 96.38 \\
        CLRNet\textbf{+Ours}~\cite{clrnet} & ResNet50~\cite{resnet} & \textbf{96.99} & 97.28 & \textbf{96.70} \\
        \hline
    \end{tabular}}
\end{table}

\begin{table*}[!t]
    \scriptsize
    \caption{Comparison with general and lane-specific SSL methods.\label{tab:ablation}}
    \centering
    \begin{tabular}{c|cccccccccc|cccc}
    
        \hline
        
        Method & F1 ($\uparrow$) & Normal & Crowded & Dazzle & Shadow & No line & Arrow & Curve & Cross & Night & Acc.($\uparrow$) & FP($\downarrow$) & FN($\downarrow$) \\
        
        \hline
        
        PixPro~\cite{pixpro} & 79.76 & 93.56 & 77.91 & \textbf{75.17} & 81.10 & 52.92 & 89.71 & 69.39 & 969 & \textbf{75.14} & 96.85 & 2.524 & 2.118 \\
        VICRegL~\cite{vicregl} & 79.19 & 93.25 & 77.42 & 73.30 & 76.90 & 53.48 & 89.84 & 69.95 & 1166 & 74.68 & 96.68 & \textbf{2.126} & 2.462 \\
        DenseCL~\cite{densecl} & 79.52 & 93.29 & 78.41 & 71.81 & 76.66 & 52.09 & 89.91 & 70.07 & 977 & 74.81 & 96.86 & 2.702 & 1.845 \\
        MoCo-V2~\cite{mocov2} & 79.70 & 93.47 & 78.68 & 72.47 & 79.19 & 52.72 & 89.96 & 68.29 & 1150 & 74.99 & 96.77 & 2.694 & 2.432 \\
        SparK~\cite{spark} & 78.77 & 92.91 & 77.23 & 70.37 & 79.09 & 50.67 & 89.34 & 65.87 & \textbf{883} & 73.68 & 96.74 & 2.719 & 2.049 \\
        CLLD~\cite{clld} & 79.27 & 92.94 & 77.44 & 72.46 & \textbf{81.20} & 53.30 & 89.31 & 68.40 & 1026 & 74.43 & 94.25 & 21.40 & 6.900 \\
        
        \hline
        
        Ours & \textbf{80.02} & \textbf{93.75} & \textbf{78.77} & 72.84 & 80.75 & \textbf{53.88} & \textbf{90.07} & \textbf{70.74} & 1507 & 75.02 & \textbf{96.89} & 2.84 & \textbf{1.57} \\
        \hline
    \end{tabular}
\end{table*}
\vspace{-0.5em}


\begin{table}[!t]
    \scriptsize
    \caption{Subset‑specific metrics from Table\ref{tab:table1}.}\label{tab:major_gains}
    \centering
    \begin{tabular}{@{\hskip 0.020in}l@{\hskip 0.020in}|
    @{\hskip 0.020in}c@{\hskip 0.020in}
    @{\hskip 0.020in}c@{\hskip 0.020in}|
    @{\hskip 0.020in}c@{\hskip 0.020in}
    @{\hskip 0.020in}c@{\hskip 0.020in}}
    
    \hline
    Method & Backbone & Night ($\uparrow$) & Backbone & Shadow ($\uparrow$) \\
    \hline
    RESA~\cite{resa} & ResNet34~\cite{resnet} & 72.80 $\rightarrow$ 75.31 & ResNet50~\cite{resnet} & 72.80 $\rightarrow$ 75.31 \\
    CLRNet~\cite{clrnet} & ResNet34~\cite{resnet} & 75.02 $\rightarrow$ 75.26 & ResNet34~\cite{resnet} & 79.92 $\rightarrow$ 80.71 \\
    CLRerNet~\cite{clrernet} & DLA34~\cite{dla34} & 76.92 $\rightarrow$ 77.07 & ResNet50~\cite{resnet} & 83.67 $\rightarrow$ 84.39\\
    \hline
    \end{tabular}
\end{table}

    
        

\begin{table*}[!t]
    \scriptsize
    \caption{Comparison with Knowledge Distillation methods.\label{tab:ablation_kd}}
    \centering
    \begin{tabular}{c|cccccccccc}
        \hline
        Method & F1 ($\uparrow$) & Normal & Crowded & Dazzle & Shadow & No line & Arrow & Curve & Cross & Night \\
        \hline
        Baseline & 81.43 & 94.36 & 80.62 & 75.23 & 84.35 & 57.31 & 91.17 & 79.11 & 1540 & 76.92\\
        KD & 81.24 & 94.15 & 80.23 & 75.31 & 84.22 & 57.25 & 90.78 & 79.17 & 1321 & 79.92\\
        SCKD & 81.51 & 94.35 & 80.52 & 75.53 & 84.55 & 56.91 & 91.00 & 78.62 & 1411 & 76.95\\
        \hline
        Ours & 81.64 & 94.42 & 80.81 & 75.83 & 84.48 & 57.55 & 90.57 & 80.25 & 1434 & 77.07\\
        \hline
    \end{tabular}
\end{table*}


    
    
    
    
    
    
    

\subsection{Ablation Study}
To evaluate the effectiveness of the proposed SAMIRO, we follow the SSL models used in the CLLD~\cite{clld} ablation study, including PixPro~\cite{pixpro}, VICRegL~\cite{vicregl}, and DenseCL~\cite{densecl}, MoCo-V2~\cite{mocov2}, on the CULane and TuSimple datasets. All experiments are conducted using ResNet50 backbone~\cite{resnet} and CLRNet~\cite{clrnet} as the lane detection framework.
SAMIRO achieves the highest overall performance with an F1 score of 80.02, surpassing general-purpose SSL baselines such as PixPro~\cite{pixpro} (79.76), DenseCL~\cite{densecl} (79.52), and MoCo-V2~\cite{mocov2} (79.70). Notably, it also outperforms the oracle model Spark~\cite{spark} (78.77) under simple transfer learning, demonstrating its ability to extract domain-agnostic features more suitable for lane detection.
SAMIRO exceeds the performance of CLLD~\cite{clld}, a Lane Detection SSL method, which reports an F1 score of 79.27. As shown in Table~\ref{tab:ablation}, SAMIRO also achieves the lowest false negative rate (1.57), indicating improved reliability in detecting all lanes.MI‑based transfer preserves spatially continuous lane cues from the oracle, whereas instance‑discrimination‑style contrastive objectives may underemphasize such structure.

Lane detection poses unique challenges in autonomous driving. Unlike object detection or depth estimation, it requires understanding the continuous geometric structure of lanes. Thus, SSL methods from other domains are not directly applicable and must be adapted to capture these specific properties.

Unlike CLLD~\cite{clld}, which requires both pretraining and fine-tuning for each architecture, SAMIRO requires only a single fine-tuning step on any downstream model if a pre-trained oracle is available. This architecture-agnostic design improves flexibility and scalability for real-world deployment. Moreover, SAMIRO adds no additional test-time computational overhead, maintaining the same GFLOPs and FPS as the baseline, which means it adds zero parameters/ops.

Table~\ref{tab:major_gains} reports performance under challenging conditions (Night and Shadow). Overall, fine-tuned models consistently outperformed baselines, showing that architectural refinements enhance robustness. CLReRNet with DLA34 achieved the best gain at night (from 76.92 to 77.07), while CLReRNet with ResNet50 showed the largest improvement under shadows (from 83.67 to 84.39).
Backbone choice also influenced resilience: ResNet34 yielded modest gains, whereas ResNet50 and DLA backbones were more effective in complex scenarios, suggesting that backbone complexity mitigates condition-specific degradation.
%

We evaluated knowledge distillation baselines using the CULane dataset and the CLRerNet~\cite{clrernet} (DLA34~\cite{dla34}) model in Table~\ref{tab:ablation_kd}. The results showed that KD~\cite{kd1} decreased the F1 score relative to the baseline, and  SCKD~\cite{kd2} produced only a small gain in performance. In contrast, SAMIRO achieved the highest F1 score, demonstrating superior performance. This supports our claim that selectively transferring lane-relevant spatial information via MI regularization is more effective than imitating logits or indiscriminately matching features.

We conducted an ablation study on the CLRerNet~\cite{clrernet} model with a ResNet50~\cite{resnet} backbone.
Using only the Loss\_samiro component achieved an F1 score of 81.36\%. Adding the Norm component increased it to 81.43\%, and incorporating all components yielded the highest score of 81.45\%, confirming the effectiveness of each part of the proposed loss function.
%

\vspace{-1.0em}

\section{Conclusion}
%
In this paper, we propose SAMIRO, a novel regularization method that functions to address the generalization challenge in lane detection.
SAMIRO regularization addresses the overfitting problem that occurs when the source and target data have different properties in a knowledge transfer scenario.
%
By applying SAMIRO on the Lane detection models~\cite{resa, ufsa, clrnet, clrernet, bezier, laneatt}, we demonstrated its effectiveness in transferring domain-agnostic knowledge from the pre-trained model to the target data.
We achieved state-of-the-art performance in lane detection by applying SAMIRO to major benchmark lane detection datasets.
%
Since the proposed method entails a low computational burden, it can be employed in a plug-and-play manner without any degradation in inference time.
We expect SAMIRO to play an important role in the various occlusion environments in automobile applications.

\section*{Acknowledgments}
This work was supported by Institute of Information \& Communications Technology Planning \& Evaluation(IITP) grant (IITP-2024-RS-2024-00417958; Global Research Support Program in the Digital Field), and the National ResearchFoundation of Korea(NRF) grant (No.RS-2023-00212484; xAI for Motion Prediction in Complex, Real-World Driving Environment) funded by the Korea government (MSIT).

\bibliographystyle{elsarticle-num} 
\bibliography{ref}








\end{document}